# Online Signature Verification Based on Writer Specific Feature Selection and Fuzzy Similarity Measure

Chandra Sekhar V
IIIT Sricity
Sricity, India
Chandrasekhar.v@iiits.in

Prerana Mukherjee
IIIT Sricity
Sricity, India
prerana.m@iiits.in

Guru D S
Mysore University
Mysore, India
dsg@compsci.unimysore.ac.in

Viswanath Pulabaigari
IIIT Sricity
Sricity, India
viswanath.p@ieee.org

## Abstract

*Online Signature Verification (OSV) is a widely used biometric attribute for user behavioral characteristic verification in digital forensics. In this manuscript, owing to large intra-individual variability, a novel method for OSV based on an interval symbolic representation and a fuzzy similarity measure grounded on writer specific parameter selection is proposed. The two parameters, namely, writer specific acceptance threshold and optimal feature set to be used for authenticating the writer are selected based on minimum equal error rate (EER) attained during parameter fixation phase using the training signature samples. This is in variation to current techniques for OSV, which are primarily writer independent, in which a common set of features and acceptance threshold are chosen. To prove the robustness of our system, we have exhaustively assessed our system with four standard datasets i.e. MCYT-100 (DB1), MCYT-330 (DB2), SUSIG-Visual corpus and SVC-2004-Task2. Experimental outcome confirms the effectiveness of fuzzy similarity metric-based writer dependent parameter selection for OSV by achieving a lower error rate as compared to many recent and state-of the art OSV models.*

1. Introduction

   Computer forensics deals with the area of of accumulate, inspect, and details on digital data so that is both legal and acceptable in court. It is an evidence-based procedure that can be used for the recognising and prevention of a cybercrime or any event that involves the misuse of digital data.

   Due to wide spread usage of digital applications through light weight mobile devices, the demand for these applications combined with the exponential growth in online applications and mobile devices usage motivates the need for research in online signature based digital forensics [1,4,5].

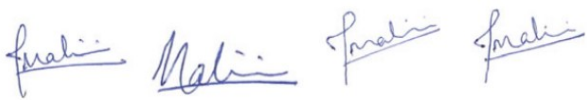

Figure 1. Types of Forgeries in case of offline signatures.(Source.[38]).

Online signature is defined by signals changing over time, which are acquired using Stylus Pens, Graphic Tablets, PCs and Smart Phones which enables reading both shape information (x, y co-ordinates) and dynamic properties (such as velocity, pressure, acceleration, azimuth, total signature time etc.,) [1,2,9,32].

Challenges for online signature verification framework are introduced by factors such as intra-individual variability (between genuine and genuine), inter-individual variability (between genuine and forgery) and requirement of high computation capabilities [1,4-6,30]. Among these variations, intra-individual variability and ability to work on computationally lightweight devices are the most challenging co-variates of online signature verification. Hence, it is imperative and challenging to enable OSV frameworks to cater to these variations [8,29,30].

In literature many techniques toward automatic on-line signature verification (OSV) have been proposed which can be broadly classified into feature-based methods [1-8,10-12,16-18,21] that analyze signatures based on a set of global or local features, function-based methods which employ various techniques like Hidden Markov models [9], sequence matching [14], Divergence based [14], DTW [1,8,18,29], Gaussian Mixture Models [23], Stability based [1,21,26], feature weighing based [20], matching based [14], Neural network based [22], Deep learning based [29] etc. In literature, we can comprehend the application of different classifiers for online signature, such as interval valued classifier [5,7,10,17], random forest [28], feature fusion based [1,7], distance or similarity based [23], SVM [32], PCA [20], Critical segments [41] and Edit distance [40] etc.

In literature, very few attempts have been done in which writer specific features and parameters are computed for OSV to efficiently preserve the characteristics of the respective writer [10,17]. Manjunatha et al [17] proposed an OSV model based on multi cluster feature selection to find a feature subset with size 'd' which contains the most informative features based on spectral embedding. Guru et al [10] had proposed a model based on writer dependent features which are selected based on top eigenvectors of graph Laplacian, computed for each feature of the respective writer.

In recent years, advances in machine learning and deep learning technologies result in evolution of Convolutional Neural Networks (CNN) based OSV frameworks [33,34,36].

Despite the lower error rates, CNN frameworks require a relatively large number of training samples for each registered user to learn the intra-individual variability, inter-individual variability [1,5,6,30] to efficiently classify the genuineness of signatures [13,30]. However, it is often impractical to obtain adequate signature samples from users, given the sensitivity of applications e.g., m-banking and m-payment [17,19,31].

Very few works explored the possibility of OSV systems with few shot learning i.e., learning the user specific features with one/few signature samples. Galbally et al [35] proposed an OSV framework in which synthetic samples are generated from one signature sample by duplicating the signature using Hidden Markov Models. Another work in the same direction is by Diaz et al [37], in which single samplings were duplicated based on the kinematic theory of rapid human movements, and its sigma-lognormal parameters. This model achieved an Equal Error rate (EER) (The point at which False Acceptance Ratio (FAR) equals False Rejection Rate (FRR)) of 13.56%.

In digital forensics, the OSV framework must work with fewer training samples. Hence, few shot learning is a critical requirement for digital forensics. Hence, in this work we focus on computationally efficient and few shot learning based OSV framework.

The manuscript is organized as follows. In Section 2, we present different phases of our proposed framework. In section 3, details of training and testing data, experimental analysis along with the results produced by the model are discussed. Computational complexity is discussed in section 4. Conclusions are drawn in section 5.

2. Proposed Online Signature Verification Model

The proposed model is divided into training, validation and testing phases.

2.1 Training phase:

In this phase a part of genuine signature samples of each user (say 'j') is split into training and validation samples. In case of MCYT dataset, out of a total of 25 genuine signatures ($GS_j$) for each user, 10 genuine signatures ($GS_{jtr}$) are used for training and 10 are used for validation ($GS_{jv}$) (for fine-tuning the parameters). $GS_{jtr}$ is used to compute the writer

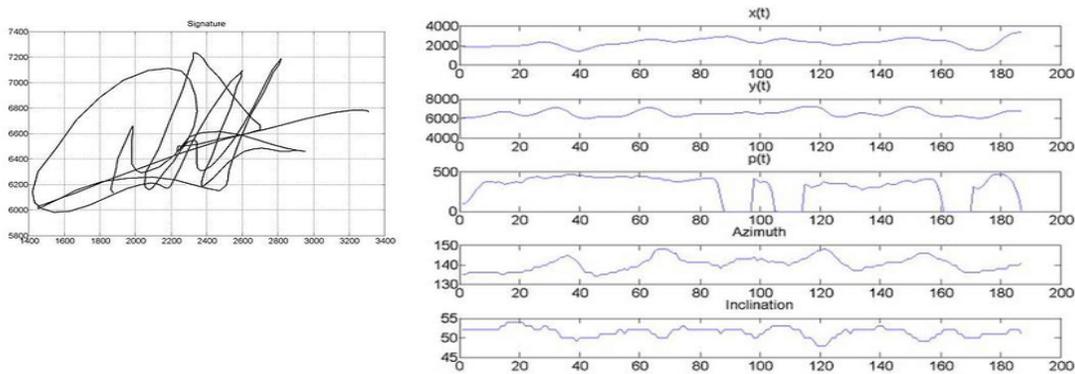

Figure 2. A sample online signature from the MCYT-100 signature corpus [39].

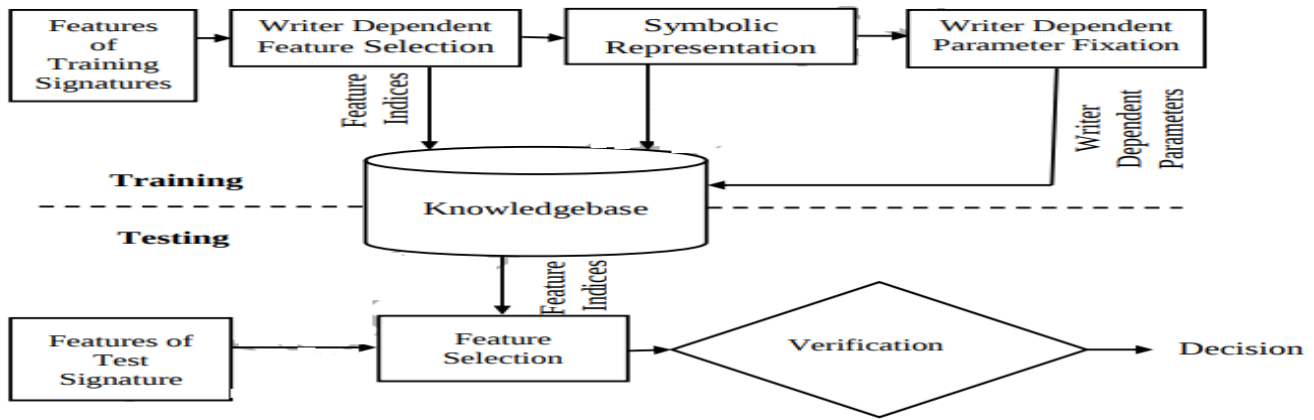

Figure 3. Illustrates the details of proposed writer specific parameter fixation framework.

Specific features. In this phase writer specific feature subset selection (selecting 80 best features out of total 100 in case of MCYT dataset) is done in three stages: 1) Computing the Median of Medians (MoM) statistical dispersion measure for each feature vector. 2) The vector containing the MoM values are given as input to the DBSCAN clustering algorithm, such that the features with relative MoMs are grouped together. DBSCAN is having intrinsic advantage of determining the number of clusters into which the users feature vector can be clustered without the information on of possible number of clusters 'K' as an input.

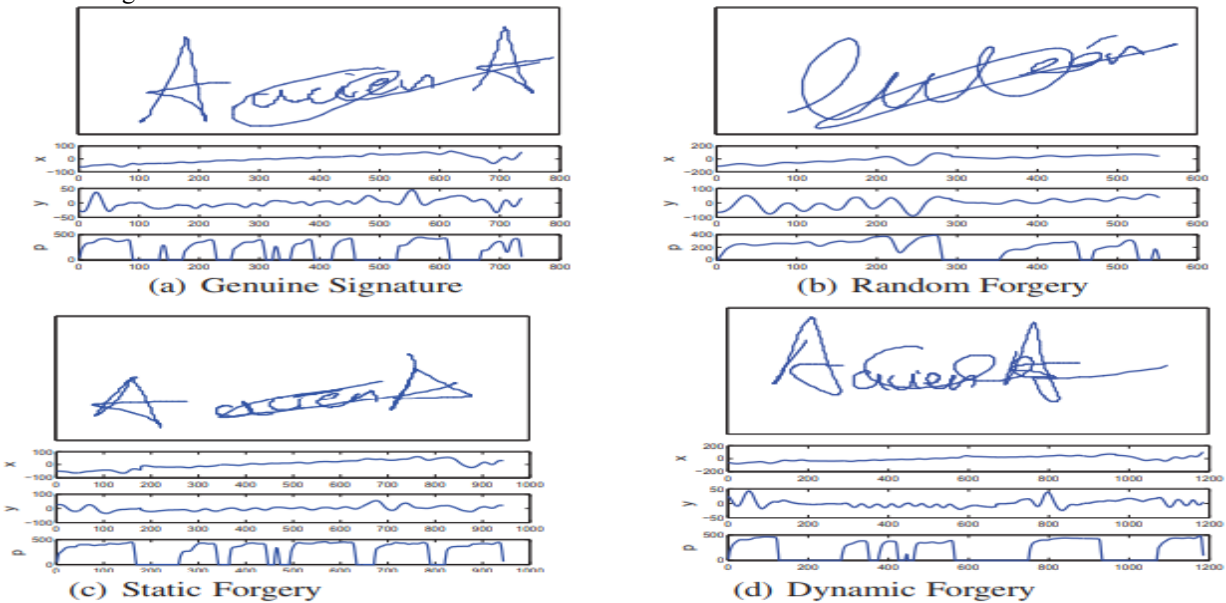

Figure 4. Types of Forgeries in case of online signatures.(Source.[38]).

The cluster having the largest number of features are selected and rest are ignored. 3) for further refining of the feature selection, the weights are computed for each feature of the selected cluster using Intelligent Minkowski k-means (imwk-means) metric [24]. Finally, the top 80% of features are selected as writer specific features (FS) and all the operations are performed on signatures with reduced feature set FS rather then full feature set. We will discuss each step in detail below:

### 2.1.1 Writer Dependent Feature selection through MoM and Clustering.

Let us consider a signature-feature matrix (SF) of writer $U_j$, with 'n' number of rows and 'm' number of columns. Each row corresponds to a signature sample of $U_j$, where j = 1, 2, 3,….N. (N represents the number of writers). And each column corresponds to a feature. Let $S^j = [S_1^j, S_2^j, S_3^j, ..., S_n^j]$ be a set of 'n' signature samples of writer 'j' i.e. $U_j$. Let $F^j = [F_1^j, F_2^j, F_3^j, ..., F_m^j]$ be a set of m-dimensional combined feature vectors, where $F_i^j = [f_{i1}^j, f_{i2}^j, f_{i3}^j, ..., f_{in}^j]$ are the feature set characterizing the $i^{th}$ feature of all the signature samples of writer 'j'.

The fundamental idea of the proposed OSV framework is to group/cluster similar features. To accomplish this, Median of Medians ($MoM_n$) is computed for each feature vector $F_i^j$. The reason for selecting MoM over other measures like Mean Absolute Difference (MAD) is that instead of measuring how far away the observations are from a central value, MoM looks at a typical distance between observations, which is effective at asymmetric distributions (Gaussian etc.).

$$\text{MoM}_n = c \ med_i\{med_j|x_i - x_j|\} \quad (1)$$

$MoM_n$ can be read as follows, for each 'i', first compute the median of $\{|x_i - x_j|; j = 1,2, ...n\}$. This results in 'n' numbers, the median of which provides the final estimate MoM. OSV suffers from two critical challenges i.e inter and intra variability. 1. The proposed interval valued symbolic representation learns intra-class variability of the signatures in each class. The usage of fuzzy similarity measure can grasp inter-class variability between genuine signatures and skilled forgeries, in which lower values of membership are provided for forged signature features compared to the genuine signature features.

### 2.1.2 Computing the user-specific features based on feature weighing.

Intelligent Minkowski k-means (imwk-means) which represents the weighted version of the Minkowski distance [24] has been used to compute the feature weight. A feature weight '$W_{kv}$' represents the degree of relevance of a feature 'v' at a cluster 'k' and is computed using squared Euclidean distance i.e. p=2.

$$D_{kv} = \sum_{u \in V} W_{kv}^P |y_{iv} - C_{kv}|^p \quad (2)$$

$$W_{kv} = \sum_{u \in V} [\frac{D_{kv}}{D_{ku}}]^{\frac{1}{p-1}^{-1}} \quad (3)$$

The procedure is repeated for twenty trials and the feature weights over the trials are averaged. Finally, the top 80% of features are selected as writer specific features (FS)

2.2 Validation phase (parameter fixation):

The validation samples ($GS_{jv}$) are used for parameter fixation. In our work, we have adopted a variant of the trapezium-shaped gaussian membership function as discussed in [23]. In our framework, two parameters need to be finetuned for each user i.e. $'\eta'$ and $'\alpha'$. $'\eta'$ describes distance from the mean value of a trapezium-shaped gaussian membership function and used in computing the interval valued representation of writer specific features. $'\alpha'$ is used in computing the writer specific acceptance threshold.

**Step 1:** As discussed above, to allow intra-user variability, for each user, we compute: $m_{jk} = mean(f_{jk}^{All})$, $s_{jk} = StdDev(f_{jk}^{All})$, where $m_{jk}$ and $s_{jk}$ represents the mean and standard deviation of feature vector 'k' of writer 'j' and the values are computed by considering all the training signature samples ($GS_j$) and features specific to writer $W_j$ computed in training phase i.e $FS_j$.

**step 2:** Compute $f_{jk}^- = m_{jk} - \eta \times s_{jk}$, $f_{jk}^+ = m_{jk} + \eta \times s_{jk}$, where $f_{jk}^-$ and $f_{jk}^+$ represent the lower and upper valid limits and 'm' and 's' represent the mean and standard deviation of the $k^{th}$ feature vector of $j^{th}$ user respectively. Similarly, all the feature vectors are represented in interval-valued form. Finally, $IVF_j = \{([f_{jk}^-, f_{jk}^+], m_{jk}, s_{jk}), FS_j\}$, where k varies from 1 to length($FS_j$) are computed.

**Step 3:** To fix the writer specific acceptance parameter, we use equation (4) and the parameter $'\alpha'$ need to be finetuned.

$$\theta_j = \text{Mean}(Fuzzy_{Sim}(W_{jtr}, IVF_j)) - \alpha_j \times StdDev(Fuzzy_{Sim}(W_{jtr}, IVF_j))$$
(4)

where $Fuzzy_{Sim}(GS_{jtr}, IVF_j)$ represents the similarity between a vector of crisp values $GS_{jtr}$ and a vector of interval valued representations $IVF_j$ where

$Fuzzy_{Sim}(GS_{jtr}, IVF_j) =$

| if $T_{jk} < f_{jk}^-$ or $T_{jk} > f_{jk}^+$ | = | 0 |
| if $(m_{jk} - \sigma_{jk}) \leq T_{jk} \leq (m_{jk} + \sigma_{jk})$ | = | 1 |
| if $f_{jk}^- \leq T_{jk} < (m_{jk} - \sigma_{jk})$ | = | $(T_{jk} - f_{jk}^+)/((m_{jk} - \sigma_{jk}) - f_{jk}^-)$ |
| if $(m_{jk} + \sigma_{jk}) < T_{jk} \leq f_{jk}^+$ | = | $(f_{jk}^+ - T_k)/(f_{jk}^+ - (m_{jk} + \sigma_{jk}))$ |

(5)

where $1 <= tr <= 10$. In each signature, 'k' varies from $1 <= k <= len(FS_j)$. $tr$ = number of training samples.

The combination of $'\eta'$ and $'\alpha'$, which results in the least Equal Error Rate (EER), the point at which the False Acceptance Ratio (FAR) and False Rejection Ratio (FRR) are equal in receiver operating characteristic (ROC) curve is finally considered. Finally, the set of values $\{IVF_j, \eta_j, \alpha_j, \theta_j\}$ specific to writer 'j' are stored in the knowledge base and will be used during the testing phase.

2.3 Testing phase:

When a test signature $T_{jtest}$ arrives, claiming that it represents writer j, $j^{th}$ writer specific details $\{IVF_j, \eta_j, \alpha_j\}$ are retrieved from the knowledge base.

a) Retrieve the writer specific feature indices from $FS_j$. Retrieve the same features from the test signature. Due to reduced features, the test signature $T_{jtest}$ becomes $T_{FSjtest}$.
b) Compute the fuzzy similarity between ($T_{FSjtest}$, $IVF_j$) using (5). If the resultant value is greater than or equal to $\theta_j$ (which represents the writer specific acceptance threshold for writer 'j'), then test signature $T_{jtest}$ is classified as genuine, else forgery.

3. Experimentation and Results

We have extensively conducted verification experiments and validated the proposed OSV framework by conducting the experiments on widely accepted four datasets i.e. MCYT_100 signature sub corpus dataset (DB1) [9,13], MCYT-330 (DB2) [6,7,10], SVC – Task 2 [12,13,18], SUSIG [8,11,13]. The results are illustrated in tables below. Table II, illustrates the indices of best writer specific features computed based on MoM statistical dispersion measure and DBSCAN clustering technique during training process (random signature category). Tables III-VI represent the comparison of EER with the latest proposed OSV frameworks, which are evaluated based on the corresponding datasets. The first best EER values are marked with (*) and the second best are marked with (**). In case of MCYT-100 (DB1) our framework achieved state-of-the-art results in S_01, S_5, S_20, R_1 and R_20 categories. In case of MCYT-100 and MCT-330, all the models considered complete 100 global features. In our framework, we have used only 80 features and still able to achieve the best state of the art results. In case of MCYT-330 (DB2) our framework achieved state-of-the-art results in all the experimentation categories. In case of SVC, the framework achieved the best EER except for R_05 and R_10 categories. In case of SUSIG, the framework achieved the best EER in all the categories.

From the tables III-VI, we confirm that in all categories and in all the datasets, the framework achieves decreasing EER with the increasing of the number of training samples. Any deviation is possibly due to false assignment of fuzzy values. Also, the tables III-VI and in Fig 5, concludes that, in both skilled categories, with the increase of training samples, SUSIG shows steep decrease in EER value,

TABLE I. THE DATASET DETAILS USED IN THE EXPERIMENTS FOR THE PROPOSED FRAMEWORK

| DataSet → | MCYT-100 | MCYT-330 | SVC | SUSIG |
|---|---|---|---|---|
| Total number of Users | 100 | 100 | 40 | 94 |
| Total number of features | 100 | 100 | 47 | 47 |
| Number of genuine signatures per user | 25 | 25 | 20 | 20 |
| Number of forgery signatures per user | 25 | 25 | 20 | 10 |
| Total number of genuine signatures | 2500 | 8250 | 800 | 1880 |
| Total number of forgery signatures | 2500 | 8250 | 800 | 940 |
| Total Number of Samples | 5000 | 16500 | 1600 | 2820 |

TABLE II. THE MOST RELEVANT 80 FEATURES WHICH BEST REPRESENTS THE WRITERS SIGNATURE (MCYT-DB1).

| Writer Id | The best feature indices to represent writers' signature. |
|---|---|
| 1 | [49;48;53;56;51;46;47;75;70;79;58;59;63;45;73;26;60;40;29;17;34;61;57;43;64;65;55;52;50;71;72;74;42;32;14;80;35;54;18;41; 19;33;28;66;23;38;36;22;44;27;30;62;13;37;16;24;15;77;78;67;25;12;31;8;76;11;39;21;20;9;10;1;2;34;5;6;7;68;69] |
| 96 | [47;49;45;50;48;73;54;46;52;75;35;43;64;66;26;32;29;18;57;77;14;78;19;65;61;15;51;63;42;41;56;80;69;74;34;13;28;24;53;17; 79;23;55;16;58;27;59;70;22;60;67;40;68;37;62;36;72;31;38;71;30;25;44;11;33;76;12;21;39;20;10;1;2;3;4;5;6;7;8;9;] |

TABLE III. COMPARATIVE ANALYSIS EER OF THE PROPOSED MODEL AGAINST THE CONTEMPORARY MODELS ON MCYT (DB1) DATABASE

| Method | S_01 | S_05 | S_20 | R_01 | R_05 | R_20 |
|---|---|---|---|---|---|---|
| **Proposed Model – (Writer specific features + Fuzzy Similarity) (80 features)** | **0.083*** | **0.025*** | **0.000*** | **0.033*** | **0.032** | **0.1100*** |
| With writer dependent parameters (Symbolic) [1] | - | 2.2 | 0.6 | - | 1.0 | 0.1 |
| Cancelable templates - HMM Protected [9] | - | 10.29 | - | - | - | - |
| Cancelable templates - HMM[9] | - | 13.30 | - | - | - | - |
| Stroke-Wise [13] | 13.72 | - | - | 5.04 | - | - |
| Target-Wise [13] | 13.56** | - | - | 4.04** | - | - |
| Information Divergence-Based Matching [14] | - | 3.16 | - | - | - | - |
| WP+BL DTW[18] | - | 2.76 | - | - | - | - |
| Combinational Features and Secure KNN-Combined features [21] | - | 3.69 | - | - | 1.08 | - |
| Stability Modulated Dynamic Time Warping (F12) [21] | - | 7.76 | - | - | 0.75 | - |
| Stability Modulated Dynamic Time Warping (F13) [21] | - | 13.56 | - | - | 4.31 | - |
| Dynamic Time Warping-Normalization(F13) [21] | - | 8.36 | - | - | 6.25 | - |
| Stability Modulated Dynamic Time Warping (F13) [21] | - | 3.09 | - | - | 1.30 | - |
| Histogram + Manhattan [25] | - | 4.02 | - | - | 1.15 | - |
| discriminative feature vector + several histograms [25] | - | 4.02 | 2.72 | - | 1.15 | 0.35 |
| VQ+DTW[27] | - | 1.55 | - | - | - | - |
| GMM+DTW with Fusion | - | 3.05 | - | - | - | - |
| RPDTW[29] | - | - | - | - | - | - |
| Probabilistic-DTW(case 1) [30] | - | - | - | - | 0.0118* | - |
| Probabilistic-DTW(case 2) [30] | - | - | - | - | 0.0187** | - |
| String Edit Distance [40] | - | 1.65 | - | - | 4.20 | - |

TABLE IV. COMPARATIVE ANALYSIS OF THE PROPOSED MODEL AGAINST THE CONTEMPORARY MODELS ON MCYT (DB2) DATABASE

| Method | S_01 | S_05 | S_20 | R_01 | R_05 | R_20 |
|---|---|---|---|---|---|---|
| **Proposed Model – (Writer specific features + Fuzzy Similarity) (80 features)** | **0.151*** | **0.023*** | **0.00*** | **0.015*** | **0.0170*** | **0.088*** |
| Signature-Legibility + Multi Layer Perceptron [6] | - | - | - | - | 0.2** | - |
| Symbolic Representation - Writer specific [7] | - | 5.96** | 4.70 | - | 1.88 | 1.67 |
| Symbolic representation - Common Threshold [7] | - | 6.45 | 5.55 | - | 2.10 | 2.16 |
| User dependent features [10] | - | 15.90 | 6.10 | - | 1.90 | 1.80 |
| writer dependent features and classifiers [17] | - | 18.41 | 0.94** | - | 7.54 | 0.67** |

TABLE V. COMPARATIVE ANALYSIS OF THE PROPOSED MODEL AGAINST THE CONTEMPORARY MODELS ON SVC DATASET

| Method | S_01 | S_05 | S_10 | R_01 | R_05 | R_10 |
|---|---|---|---|---|---|---|
| **Proposed Model – (Writer specific features + Fuzzy Similarity) (40 features)** | **0.197*** | **0.083*** | **0.125*** | **0.902*** | **0.904** | **0.916** |
| LCSS (User Threshold) [12] | - | - | 5.33 | - | - | - |
| Target-Wise [13] | 18.63 | - | - | 0.50* | | |
| Stroke-Wise [13] | 18.25** | - | - | 1.90 | - | - |
| DTW based (Common Threshold) [18] | - | - | 7.80 | - | - | - |
| Stroke Point Warping [19] | | - | 1.00** | | - | - |
| SPW+mRMR+SVM(10-Samples) [19] | - | - | 1.00** | - | - | - |
| Variance selection [20] | - | - | 13.75 | - | - | - |
| PCA [20] | - | - | 7.05 | - | - | - |
| Relief-1 (using the combined features set) [20] | - | - | 8.1 | - | - | - |
| Relief-2 [20] | - | - | 5.31 | - | - | - |
| RNN+LNPS[29] | - | - | - | - | 2.37 | - |
| Probabilistic-DTW(case 1) [30] | - | - | - | - | 0.0025* | - |
| Probabilistic-DTW(case 2) [30] | - | - | - | - | 0.0175** | - |

TABLE VI. COMPARATIVE ANALYSIS OF THE PROPOSED MODEL AGAINST THE CONTEMPORARY MODELS ON SUSIG DATASET

| Method | S_01 | S_05 | S_10 | R_01 | R_05 | R_10 | Number of Samples for training |
|---|---|---|---|---|---|---|---|
| **Proposed Model – (Writer specific features + Fuzzy Similarity) (40 features)** | 0.7580* | 0.7560* | 0.790** | 0.106* | 0.095* | 0.08* | |
| cosα, speed + enhanced DTW [8] | - | - | 3.06 | - | - | - | 10 |
| pole-zero models [11] | - | - | 2.09 | - | - | - | 05 |
| Target-Wise [13] | 6.67** | | | 1.55** | | | |
| Stroke-Wise [13] | 7.74 | | | 2.23 | | | |
| Information Divergence-Based Matching [14] | - | 1.6** | 2.13 | - | - | - | - |
| DCT and sparse representation [15] | - | - | 0.51* | - | - | - | 05 |
| with all domain [16] | - | - | 3.88 | - | - | - | 10 |
| with stable domain [16] | - | - | 2.13 | - | - | - | 10 |
| Length Normalization + Fractional Distance [22] | - | - | 3.52 | - | - | - | 10 |
| String Edit Distance [40] | - | 1.70 | - | - | - | 2.91 | 05 |

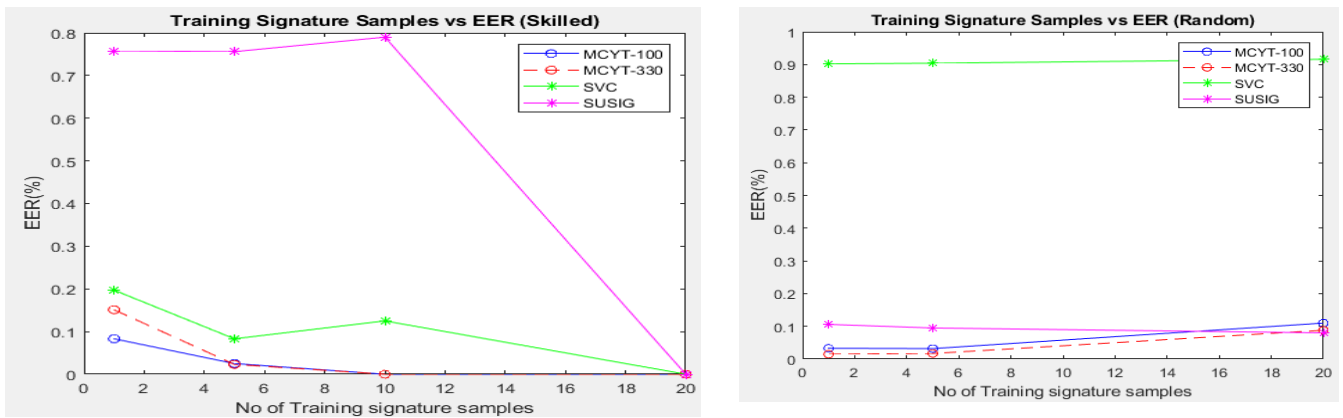

Figure. 5. The average EER with four different datasets for Skilled Forgeries (a) and for Random Forgeries (b).

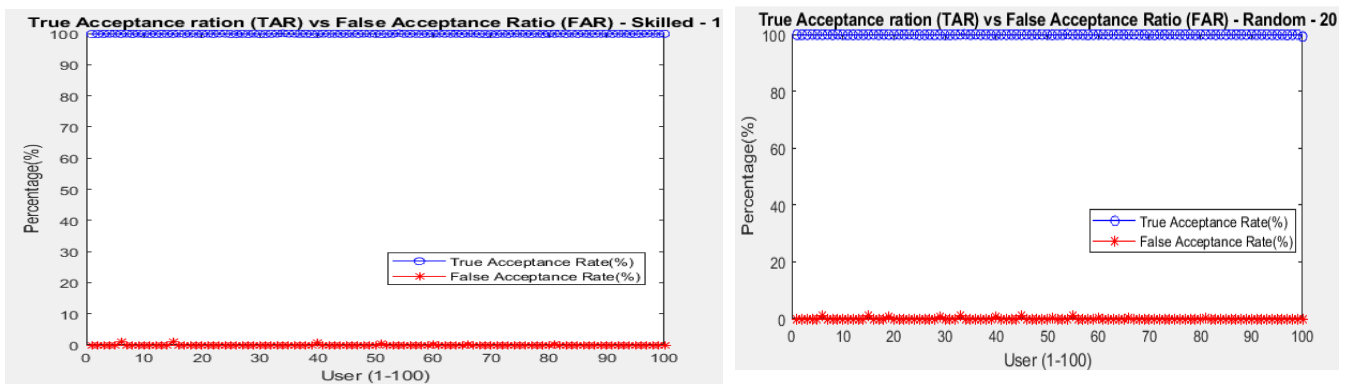

Figure 6. (a) the TAR and FAR for 100 users with 1 training samples for each user under Skilled 1 category. (b) the TAR and FAR for 100 users with 1 training samples for each user under Random 20 category of MCYT-100 dataset.

followed by MCYT-330 and MCYT-100. Figure 3 represents the 2D-Histogram of EER of each user in Skilled 5 category of SUSIG dataset and Random 1 category of SVC dataset. As illustrated in table III-VI, even though the framework proposed in [30] are resulting in better EER values compared to the proposed framework, In case of MCYT-100 (DB1) and SVC datasets, our model out performs the state of the models except the recently proposed probabilistic-DTW based OSV model by Al-Hmouz et al [30]. The computational complexity of our proposed model is $O(d)$ whereas Al-Hmouz et al [30] is $O(d^2)$ where '$d$' is the dimension of the feature vector.

Also, Al-Hmouz et al model is not extensively evaluated with other datasets (MCYT-330, SUSIG) and all categories of training like skilled_1, random_1 etc. these models are not extensively evaluated with categories of skilled_1, random_1, whereas we have evaluated the model with all the possible training samples (1,5,10,15,20) and the performance is evaluated.

Fig 6, illustrates the Receiver Operating Characteristic (ROC) curves for 100 users of Skilled_1 and Random_20 categories of MCYT-100 database. Fig 5a depicts the TAR and FAR for 100 users of MCYT-100 dataset. As the framework is trained with one genuine and one forgery signature sample of each user and tested with 24-Genuines and 24-Forgery signature samples, the TAR varies between 97-100% and FAR varies between 0-0.5%. In case of Random 20 category, as the framework is trained with 1 genuine signature and 99 genuine of other users, the TAR varies between 93-100% and FAR is recorded 0-0.5%. Hence, we confirm that the framework reflects the realistic scenario.

4. Computational Complexity Analysis of Proposed Model

The two critical steps in our work are signature enrollment and signature verification. We have considered only the verification stage, as enrollment is a one-time, offline activity during the model training.

TABLE VII.    COMPARATIVE ANALYSIS OF THE COMPUTATIONAL COMPLEXITY

| Method | Time Complexity |
| --- | --- |
| HMM-based algorithms [18] | $O(g^2 d)$ |
| GMM-based approach [23] | $O(pgd^3)$ |
| NN-based approach [22] | $O(2^d)$ |
| SVM-based methods [32] | $O(d^3)$ |
| DTW-based approach [27,28,30] | $O(d^2)$ |
| Trapezium-shaped gaussian membership [23] | $O(d)$ |
| Proposed method | $O(d)$ |

The time complexity of various methods is illustrated in Table VII. Similar to [23], verification stage is an online process, which involves computation of statistical metrics mean, standard deviation and comparison. All these operations are of linear complexity i.e. $O(d)$. '$p$' represents the data points, '$g$' is the number of Gaussian components and '$d$' is the dimension of the feature vector. Table VII confirms that the proposed Fuzzy similarity based OSV framework achieves lower EER values with less computational complexity compared to other state-of-the models.

5. Conclusion and future work

In this manuscript, we present a novel, light weight model for online signature verification grounded on user dependent feature selection and trapezium-shaped gaussian membership similarity metric. In addition, our method is computationally efficient as it works on reduced feature subset. Learning from fewer samples has been recognized as an important direction for machine learning and more so for OSV systems. The model achieved state-of-the art results in one shot learning and in various categories of all the four datasets. The proposed model has been thoroughly tested using widely accepted datasets. Experimental results demonstrate that the proposed model achieves best EER with all the four datasets. Our future work will be focusing on the development of more enriched network framework using Recurrent Neural Networks (RNN) which are best in analyzing the time series data.

**Acknowledgements.** The work reported in this paper is partly funded by a DIC funded project with reference DIC Proj14 PV.Partly from IIIT SriCity seed grant on writer recognition.